\documentclass{article}
\usepackage{hyperref}
\begin{document}

\title{Can Artificial Intelligence Do Everything That We Can?}
\author{Vincent Conitzer}
\date{}
\maketitle

\noindent {\em A
  \href{https://www.wsj.com/articles/natural-intelligence-still-has-its-advantages-1535495256}{shorter
    version of this article} appeared in {\em The Wall Street Journal}
  under the title {\em Natural Intelligence Still Has Its Advantages}, on
  August
  28, 2018~\cite{Conitzer18:Natural}.}\\

The late Stephen Hawking has warned that 
\href{http://www.bbc.com/news/technology-30290540}{AI could
  eventually ``spell the end of the human race.''}  Elon Musk has predicted
that \href{https://www.cnbc.com/2017/07/17/elon-musk-robots-will-be-able-to-do-everything-better-than-us.html}{``robots will be able to do everything better than us.''}
Meanwhile, AI systems are starting to outperform people in domains ranging
from \href{https://www.technologyreview.com/the-download/609697/deepminds-groundbreaking-alphago-zero-ai-is-now-a-versatile-gamer/}{board games}
to \href{http://www.businessinsider.com/microsoft-research-beats-humans-at-speech-transcription-2017-8}{speech recognition}.
Is humanity on the way out?

For those not working in AI, it can be difficult to interpret highly
visible achievements in the field.  Take, for example, Watson's 2011
victory over human Jeopardy champions Brad Rutter and Ken Jennings.  This
was a stunning achievement: while it should surprise nobody that Watson had
access to an encyclopedic amount of knowledge, Jeopardy is a game that
requires more than that.  The hard part -- at least for AI systems, but
often also for humans -- isn't having access to the relevant information,
but rather understanding the clue well enough to link it to that.  Even
many AI researchers, myself included, thought this would remain beyond the
capabilities of AI systems for a while to come.  We were wrong.

But does this mean that Watson had obtained a human-level understanding of
the world?  No.  Watson also produced some cringeworthy responses, for
example ``What is Toronto?'' for a clue in the ``US cities'' category.  This is
part of a broader pattern of AI systems achieving superhuman levels of
performance, and yet making blunders that leave us scratching our heads.
For example, researchers from Carnegie Mellon were able to consistently
fool a face recognition system that one of them, clearly a man, was actress
Milla Jovovich, by wearing \href{https://motherboard.vice.com/en_us/article/pgkxgv/glasses-fool-facial-recognition}{carefully designed eyeglass frames}~\cite{Sharif16:Accessorize}.

In both cases, what causes the mistake is that the AI system solves the
problem in a way that is very different from how humans do it.  Often, this
involves picking up on some statistical pattern that can be used to
surprisingly great effect, but that sometimes produces answers that lack
any common sense.  Moreover, if something changes about how the data is
produced, performance may plummet.  This is especially so when the change
is intended to mislead the system, as in the case of adding the eyeglass
frames.

This gives some insight into which jobs, or parts of jobs, the AI systems
of today and tomorrow are likely to take over from us.  Tasks that require
responding to the same kind of standardized input over and over again, with
a clear measure of success, are a natural fit.  Such tasks range from \href{https://www.cbsnews.com/news/ai-better-than-dermatologists-at-detecting-skin-cancer-study-finds/}{the
diagnosis of medical images}
to \href{https://www.cbsnews.com/news/burger-flipping-robot-flippy-starts-shift-at-caliburger/}{flipping burgers}.
On the other hand, jobs that are messy and unpredictable and require
understanding of people and the broader world -- I like to think of
kindergarten teachers -- will likely remain safe for a long time.  Driving a
car in a busy city is probably somewhere in the middle.

Much progress has been made in AI in a short span of time, so it is not
unthinkable that there will be further breakthroughs, especially if we
think in terms of decades or a century.  For now, humans remain unsurpassed
in their broad, integrated, flexible, and robust understanding of the
world.  If AI starts to catch up with us on that, it will likely change our
world beyond recognition, and some of most intractable problems in
philosophy, such as the nature of consciousness, will become very
pertinent.  I personally am interested in questions about the nature of
consciousness, and have even done \href{https://link.springer.com/article/10.1007/s10670-018-9979-6}{some research on what I consider to be its
more mysterious aspects}~\cite{Conitzer18:Puzzle}.  (How could a lump
of matter possibly give rise to {\em this} experience?)  But there is
currently no clear path towards building broadly intelligent systems.  The
AI systems we know how to build today are likely to be disruptive in many
domains -- the labor market, our social fabric, the nature of warfare.  But
they do not make humanity obsolete.

\bibliographystyle{plain}
\bibliography{/usr/project/conitzer/vccollab/references.bib}

\end{document}